\documentclass[conference]{IEEEtran}
\IEEEoverridecommandlockouts
\usepackage{amsmath}
\usepackage[utf8]{inputenc} % allow utf-8 input
\usepackage[T1]{fontenc}    % use 8-bit T1 fonts
\usepackage{hyperref}       % hyperlinks
\hypersetup{draft}
\usepackage{url}            % simple URL typesetting
\usepackage{booktabs}       % professional-quality tables
\usepackage{bbm}
\usepackage{amsfonts}       % blackboard math symbols
\usepackage{nicefrac}  
\usepackage{amsthm}% compact symbols for 1/2, etc.
\usepackage{microtype}      % microtypography
\usepackage{xcolor}         % colors
\usepackage{amsmath}
\usepackage{mathtools}
\usepackage{booktabs}
\usepackage{caption}
\usepackage{subcaption}
\usepackage[ruled,vlined]{algorithm2e}
\usepackage{graphicx}
\DeclareMathOperator*{\argmin}{arg\,min}
\newtheorem{theorem}{Theorem}

\graphicspath{ {./} }
\usepackage{caption}
\usepackage{subcaption}
\usepackage{graphicx}
\graphicspath{ {./} }

\newcommand{\norm}[1]{\left\lVert#1\right\rVert}
\def\BibTeX{{\rm B\kern-.05em{\sc i\kern-.025em b}\kern-.08em
    T\kern-.1667em\lower.7ex\hbox{E}\kern-.125emX}}
    
\newenvironment{hproof}{%
  \proof}{\endproof}
\begin{document}

\title{User-Centric Federated Learning}

\makeatletter
\newcommand{\linebreakand}{%
  \end{@IEEEauthorhalign}
  \hfill\mbox{}\par
  \mbox{}\hfill\begin{@IEEEauthorhalign}
}
\makeatother
\author{
  \IEEEauthorblockN{Mohamad Mestoukirdi $^{\dagger}$, Matteo Zecchin$^{\dagger}$, David Gesbert, Qianrui Li and Nicolas Gresset}
    \thanks{$^\dagger$ Equal contribution.\newline M. Mestoukirdi, M.  Zecchin and D. Gesbert are with the Communication
Systems Department, EURECOM, Sophia-Antipolis, France. Emails: \{mestouki, zecchin, gesbert\}@eurecom.fr.  Q.Li and N. Gresset are with Mitsubishi Electric R\&D Centre Europe. Emails: \{Q.Li, N.Gresset\}@fr.merce.mee.com. \newline The work of M. Zecchin is funded by the Marie Sklodowska Curie action WINDMILL (grant No. 813999).
\newline The  work  of  M. Mestoukirdi  is  funded  by  Mitsubishi Electric R\&D Centre Europe.}
}
\maketitle
\begin{abstract}
Data heterogeneity across participating devices poses one of the main challenges in federated learning as it has been shown to greatly hamper its convergence time and generalization capabilities. In this work, we address this limitation by enabling personalization using multiple user-centric aggregation rules at the parameter server. Our approach potentially produces a personalized model for each user at the cost of some extra downlink communication overhead. To strike a trade-off between personalization and communication efficiency, we propose a broadcast protocol that limits the number of personalized streams while retaining the essential advantages of our learning scheme. Through simulation results, our approach is shown to enjoy higher personalization capabilities, faster convergence, and better communication efficiency compared to other competing baseline solutions. %
%Thanks to the soft clustering approach followed by our solution, it is shown to over perform in scenarios where a clear clustering rule does not exist and hard clustering approaches fail.
\end{abstract}

\section{Introduction}

Federated learning \cite{mcmahan2017communication} has seen great success, being able to solve distributed learning problems in a communication-efficient and privacy-preserving manner. Specifically, federated learning provides to clients (e.g. smartphones, IoT devices, and organizations) the possibility of collaboratively train a model under the orchestration of a parameter server (PS) by iteratively aggregating locally optimized models and without off-loading local data
\cite{kairouz2019advances}. The original aggregation policy was implemented by Federated Averaging (FedAvg) \cite{mcmahan2017communication}, has been devised under the assumption that clients' local datasets are statistically identical, an assumption that is hardly met in practice. In fact, clients typically store datasets that are statistically heterogeneous and different in size \cite{sattler2020clustered}, and are mainly interested in learning models that generalize well over their local data distribution through collaboration. Generally speaking,  FedAvg exhibits slow convergence and poor generalization capabilities in such non-IID setting \cite{li2018federated}. To address these limitations, a large body of literature deals with personalization as a technique to reduce the detrimental effect of non-IID data. 
%In \cite{li2018federated} the authors augment the original FL objective function at each client with a proximal term that stabilizes the training process by limiting the divergence of the locally trained local model from the global one 
A straightforward solution consists in producing adapted models at a device scale by local fine-tuning procedures. Borrowing ideas from Model Agnostic Meta-Learning (MAML) \cite{finn2017model}, federated learning can be exploited in order to find a launch model that can be later personalized at each device using few gradient iterations \cite{fallah2020personalized,jiang2019improving}. 
Alternatively, local adaptation can be obtained by tuning only the last layer of a globally trained model \cite{arivazhagan2019federated} or by interpolating between a global model and locally trained ones \cite{deng2020adaptive,hanzely2020federated}. However, these methods can fail at producing models with an acceptable generalization performance even for synthetic datasets \cite{caldas2018leaf}.  Adaptation can also be obtained leveraging user data similarity to personalize the training procedure.
For instance, a Mixture of Experts formulation has been considered to learn a personalized mixing of the outputs of a commonly trained set of models 
\cite{reisser2021federated}. Similarly, \cite{mixture2021s} proposed a distributed Expectation-Maximization (EM) algorithm  concurrently converges to a set of shared hypotheses and a personalized linear combination of them at each device. Furthermore, \cite{zhang2020personalized} proposed a personalized aggregation rule at the user side based on the validation accuracy of the locally trained models at the different devices. In order to be applicable, these techniques need to strike a good balance between communication overhead and the amount of personalization in the system. In fact, if on one hand, the expressiveness of the mixture is proportional to the number of mixed components; on the other, the communication load is linear in this quantity.
%Multi-Task Federated Learning (MTFL) considers the minimization of the aggregate local loss terms along with a regularization term that is trained to capture similarities among clients tasks \cite{smith2017federated}. These algorithms usually employ contrived alternating optimization procedures in order to jointly learn the model and the similarities among tasks.
Clustered Federated Learning (CFL) measures the similarity among the model updates during the optimization process in order to lump together users in homogeneous groups. For example, \cite{briggs2020federated,sattler2020clustered} proposed a hierarchical strategy in which the original set of users is gradually divided into smaller groups and, for each group, the federated learning algorithm is branched in a new decoupled optimization problem. 

In this work, we propose a different approach to achieve personalization by allowing multiple user-centric aggregation strategies at the PS. The mixing strategies account for the existence of heterogeneous clients in the system and exploit estimates of the statistical similarity among clients that are obtained at the beginning of the federated learning procedure. Furthermore, the number of distinct aggregation rules --- also termed personalized streams --- can be fixed in order to strike a good trade-off between communication and learning efficiency.
%These approaches employ a hard clustering scheme that prevents the transfer of knowledge among devices in different groups; on the contrary, our approach allows each users to contribute to each different personalized model. Additionally, our algorithm performs clustering at the beginning of the federated learning process, avoiding the training overhead due to the successive refinements of hierarchical clustering. In \cite{zhao2018federated}  the accuracy reduction incurred by FedAvg in non-IID settings is quantified as a function of the Earthmover’s distance between the local distribution over classes on each client and the population class distribution. Consequently, a data sharing protocol that reduces the heterogeneity of data among the population is proposed.Our solution differ in principle from these, as they all produce a global model to be shared for the users in the same cluster, throughout the limited collaboration of devices associated to the same cluster only. Such a model is inherently adapted to the cluster averaged distribution that may still be unable to fit each user. As a result, heterogeneity may hinder the convergence time and inference accuracy over certain subpopulations even in such cases.
\subsection*{In particular, the contributions of this work are:} 

\begin{enumerate}

\item We propose a user-centric aggregation rule to personalize users' local models. This rule exploits a novel similarity score that quantifies the discrepancy between individual user data distributions. Different from previous algorithms based on user clustering $\cite{sattler2020clustered,briggs2020federated}$, our approach enables collaboration across all the nodes during training and, as a result, outperforms the above techniques, especially when clear clusters of users do not exist. Conversely to \cite{mixture2021s}, personalization is performed at the PS and therefore without the additional cost of transmitting multiple models to each client.

\item Leveraging results from domain adaptation theory, we provide an upper bound on the risk w.r.t. the local data distribution of the personalized models obtained by our aggregation strategy. The result is used to obtain insights on how to determine the degree of collaboration among the devices.
 
\item We propose a heuristic strategy to compute the mixing coefficients for the personalized aggregation without accounting for the communication overhead. Then, in order to limit the communication burden introduced by the personalized aggregation, we propose to limit the number of personalized streams using the centroids obtained by clustering the mixing coefficient vectors.

\item We provide simulation results for different scenarios and demonstrate that our approach exhibits faster convergence, higher personalization capabilities, and communication efficiency compared to other popular baseline algorithms.
\end{enumerate}
\section{Learning with heterogeneous data sources}
\label{sec3}
In this section, we provide theoretical guarantees for learners that combine data from heterogeneous data distributions. The set-up mirrors the one of personalized federated learning and the results are instrumental to derive our user-centric aggregation rule. In the following, we limit our analysis to the discrepancy distance (\ref{disc_dist}) but it can be readily extended to other divergences \cite{Journal}.

In the federated learning setting, the weighted combination of the empirical loss terms of the collaborating devices represents the customary training objective. Namely, in a distributed system with $m$ nodes, each endowed with a dataset $\mathcal{D}_i$ of $n_i$ IID samples from a local distribution $P_i$, the goal is to find a predictor $f:\mathcal{X}\to\mathcal{\hat{Y}}$ from a hypothesis class $\mathcal{F}$ that minimizes
\begin{equation}
    L(f,\vec{w})=\sum_{i=1}^m \frac{w_i}{n_i}\sum_{(x,y)\in \mathcal{D}_i}\ell(f(x),y)
    \label{aggregatedloss}
\end{equation}
where  $\ell: \mathcal{\hat{Y}}\times \mathcal{Y}\to \mathbb{R}^+$ is a loss function and $\vec{w}=(w_1,\dots,w_m)$ is a weighting scheme. In case of identically distributed local datasets, the typical weighting vector is $\vec{w}=\frac{1}{\sum_i n_i}\left(n_1,\dots,n_m\right)$, the relative fraction of data points stored at each device. This particular choice minimizes the variance of the aggregated empirical risk, which is also an unbiased estimate of the local risk at each node in this scenario. However, in the case of heterogeneous local distributions, the minimizer of $\vec{w}$-weighted risk may transfer poorly to certain devices whose target distribution differs from the mixture $P_{\vec{w}}=\sum^m_{i=1}w_iP_i$. Furthermore, it may not exists a single weighting strategy that yields a universal predictor with satisfactory performance for all participating devices. To address the above limitation of a universal model, personalized federated learning allows adapting the learned solution at each device. In order to better understand the potential benefits and drawbacks coming from the collaboration with statistically similar but not identical devices, let us consider the point of view of a generic node $i$ that has the freedom of choosing the degree of collaboration with the other devices in the distributed system. Namely, identifying the degree of collaboration between node $i$ and the rest of users by the weighting vector $\vec{w}_i=(w_{i,1},\dots,w_{i,m})$ (where $w_{i,j}$ defines how much node $i$ relies on data from user $j$), we define the personalized objective for user $i$ 
\begin{equation}
    L(f,\vec{w}_i)=\sum_{j=1}^m \frac{w_{i,j}}{n_j}\sum_{(x,y)\in \mathcal{D}_j}\ell(f(x),y)
\label{centricloss}
\end{equation}
and the resulting personalized model 
\begin{equation}
    \hat{f}_{\vec{w}_i}=\argmin_{f\in\mathcal{F}}L(f,\vec{w}_i).
    \label{learner}
\end{equation}
We now seek an answer to: \emph{``What's the proper choice of $\vec{w}_i$ in order to obtain a personalized model $\hat{f}_{\vec{w}_i}$ that performs well on the target distribution $P_i$?''}. This question is deeply tied to the problem of domain adaptation, in which the goal is to successfully aggregate multiple data sources in order to produce a model that transfers positively to a different and possibly unknown target domain. In our context, the dataset $\mathcal{D}_i$ is made of data points drawn from the target distribution $P_i$ and the other devices' datasets provide samples from the sources $\{P_j\}_{j\neq i}$.  Leveraging results from domain adaptation theory \cite{ben2010theory}, we provide learning guarantees on the performance of the personalized model $\hat{f}_{\vec{w}_i}$ to gauge the effect of collaboration that we later use to devise the weights for the user-centric aggregation rules. 

In order to avoid negative transfer, it is crucial to upper bound the performance of the predictor w.r.t. to the target task. The discrepancy distance introduced in \cite{mansour2009domain} provides a measure of similarity between learning tasks that can be used to this end. For a functional class $\mathcal{F}:\mathcal{X}\rightarrow\hat{\mathcal{Y}}$ and two distributions $P,Q$ on $\mathcal{X}$, the discrepancy distance is defined as 
\begin{equation}
    d_{\mathcal{F}}(P,Q)=\sup_{f,f'\in \mathcal{F}}\left|\mathbb{E}_{x\sim P}\left[\ell(f,f')\right]-\mathbb{E}_{x\sim Q}\left[\ell(f,f')\right]\right|
     \label{disc_dist}
\end{equation}
where we streamlined notation denoting $f(x)$ by $f$.
For bounded and symmetric loss functions that satisfy the triangular inequality,  the previous quantity allows obtaining the following inequality
\begin{equation*}
    \mathbb{E}_{(x,y)\sim P}[\ell(f,y)]\leq \mathbb{E}_{(x,y)\sim Q}[\ell(f,y)]+d_{\mathcal{F}}(P,Q)+\lambda
\end{equation*}
where $\lambda=\inf_{f \in \mathcal{F}}\left(\mathbb{E}_{(x,y)\sim P}[\ell(f,y)]+\mathbb{E}_{(x,y)\sim Q}[\ell(f,y)]\right)$. We can exploit the inequality to obtain the following risk guarantee for $\hat{f}_{\vec{w}_i}$ w.r.t the true minimzer $f^*$ of the risk for the distribution $P_i$.

\begin{theorem}
For a symmetric and $B$-bounded range loss function $\ell$ that satisfies the triangular inequality, w.p. $1-\delta$ the predictor $f_{\vec{w_i}}$  satisfies
\begin{align*}
    &\mathbb{E}_{(x,y)\sim P_i}[\ell(\hat{f}_{\vec{w}_i},y)]-\mathbb{E}_{(x,y)\sim P_i}[\ell(f^*,y)]\leq \\B &\sqrt{\sum^m_{j=1}\frac{w^2_{i,j}}{n_j}}\left(\sqrt{\frac{2d}{\sum_i n_i}\log\left(\frac{e\sum_i n_i}{d}\right)}+\sqrt{\log\left(\frac{2}{\delta}\right)}\right)\\&+2 \sum_{j=1}^m w_{i,j}d_{\mathcal{F}}(P_i,P_{j})+2\lambda
\end{align*}

where  $d$ is the VC-dimension of the function space resulting from the composition of $\mathcal{F}$ and $\ell$ and $\lambda=\argmin_{f\in \mathcal{F}}\left( \mathbb{E}_{(x,y)\sim P_i}[\ell(f,y)]+\mathbb{E}_{(x,y)\sim P_{\vec{w}_i}}[\ell(f,y)]\right)$ .
\label{Th1}

\end{theorem}
\begin{hproof}
The proof works by bounding the population risk of (\ref{learner}) w.r.t. the local measure $P_i$ and, subsequently, the estimation error of the weighted empirical risk minimizer. Full details are provided in \cite{Journal}.
\end{hproof}
The theorem highlights that a fruitful collaboration should strike a balance between the bias terms due to dissimilarity between the local distribution and the risk estimation gains provided by the data points of other nodes. Analytically minimizing the upper bounds seems an appealing solution; however, the divergence terms are difficult to compute, especially under the privacy constraints that federated learning imposes. For this reason, in the following, we consider a heuristic method based on the similarity of the readily available users' model updates to estimate the collaboration coefficients.

\section{User-centric aggregation}
\begin{figure}
         \centering
         \includegraphics[width=0.335\textwidth]{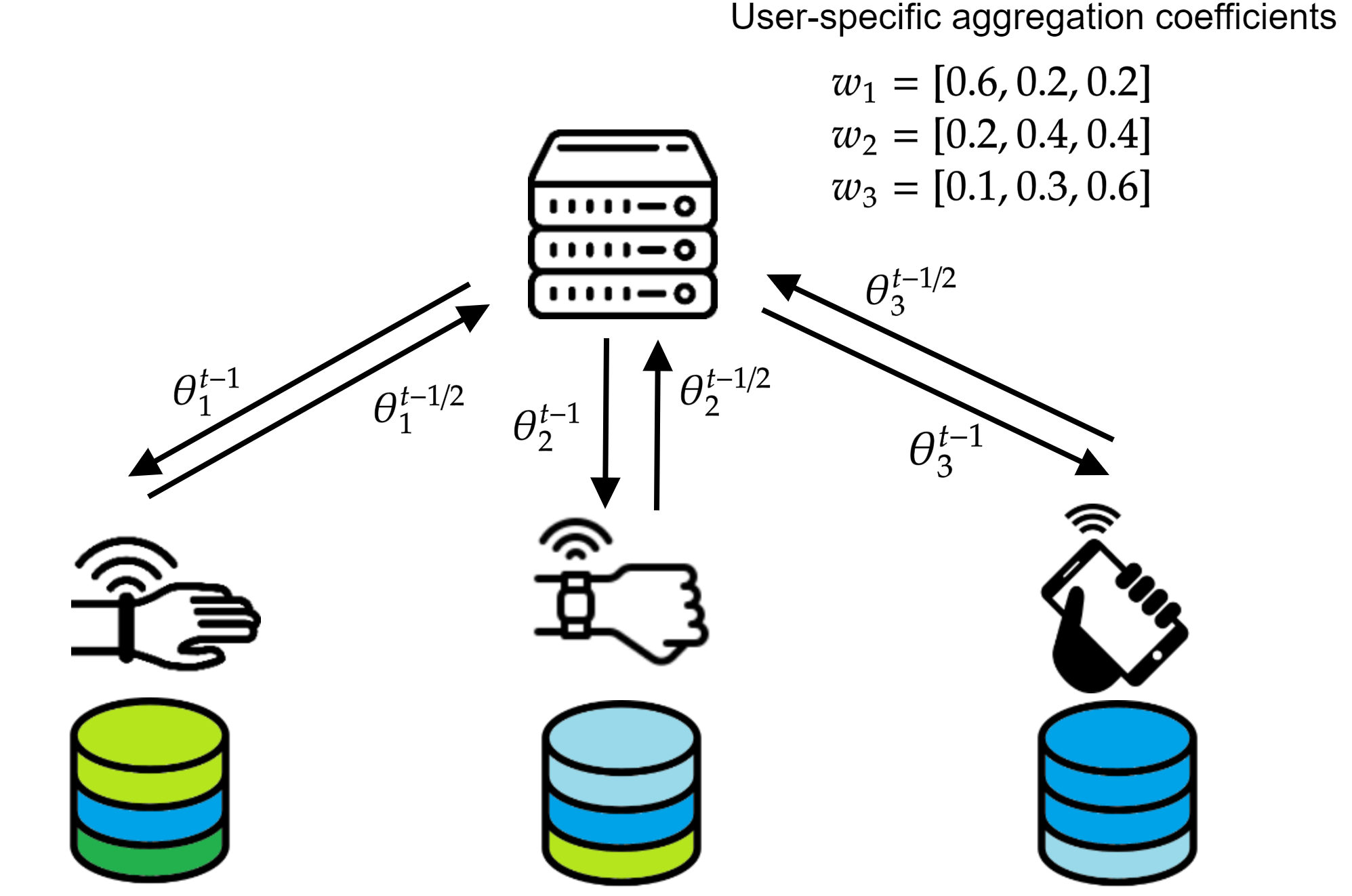}
         \label{fig:system_model}
     \caption{Personalized Federated Learning with user-centric aggregates at round $t$.}
     \label{fig:comm_eff}
\end{figure}\
For a suitable hypothesis class parametrized by $\theta\in \mathbb{R}^d$, federated learning approaches use an iterative procedure to minimize the aggregate loss (\ref{aggregatedloss}) with $\vec{w}=\frac{1}{\sum_i n_i}\left(n_1,\dots,n_m\right)$. At each round $t$, the PS broadcasts the parameter vector $\theta^{t-1}$ and then combines the locally optimized models by the clients $\{\theta_i^{t-1}\}_{i=1}^m$ according to the following aggregation rule
\[
\theta^{t}\leftarrow \sum_{i=1}^m\frac{n_i}{\sum_{j=1}^m n_j}\theta_i^{t-1}.
\]
As mentioned in Sec. \ref{sec3}, this aggregation rule has two shortcomings: it does not take into account the data heterogeneity across users, and it is bounded to produce a single solution. For this reason, we propose a user-centric model aggregation scheme that takes into account the data heterogeneity across the different nodes participating in training and aims at neutralizing the bias induced by a universal model. Our proposal generalizes the naïve aggregation of FedAvg, by assigning a unique set of mixing coefficients $\vec{w}_i$ to each user $i$, and consequently, a user-specific model aggregation at the PS side. Namely, at the PS side, the following set of user-centric aggregation steps are performed
\begin{equation}
\begin{aligned}
\theta^{{t}}_i \leftarrow \sum^m_{j=1}w_{i,j}\theta^{{t-1/2}}_j \quad \textnormal{for $i=1,\dots,m$}
\end{aligned}
\label{eq3}
\end{equation}
 where now, $\theta^{{t-1/2}}_j$ is the locally optimized model at node $j$ starting from $\theta^{{t-1}}_j$, and $\theta^{{t}}_i$ is the user-centric aggregated model for user $i$ at communication round $t$.

As we elaborate next, the mixing coefficients are heuristically defined based on a distribution similarity metric and the dataset size ratios. These coefficients are calculated before the start of federated training. The similarity score we propose is designed to favor collaboration among similar users and takes into account the relative dataset sizes, as more intelligence can be harvested from clients with larger data availability. Using these user-centric aggregation rules, each node ends up with its own personalized model that yields better generalization for the local data distribution. It is worth noting that the user-centric aggregation rule does not produce a minimizer of the user-centric aggregate loss given by (\ref{centricloss}). At each round, the PS aggregates model updates computed starting from a different set of parameters. Nonetheless, we find it to be a good approximation of the true update since personalized models for similar data sources tend to propagate in a close neighborhood. The aggregation in \cite{zhang2020personalized} capitalizes on the same intuition.
\subsection{Computing the collaboration coefficients}
Computing the discrepancy distance (\ref{disc_dist}) can be challenging in high-dimension, especially under the communication and privacy constraints imposed by federated learning. For this reason, we propose to compute the mixing coefficient based on the relative dataset sizes and the distribution similarity metric given by
\begin{align*}
\Delta_{i,j}(\hat{\theta}) = & \norm{ \frac{1}{n_i}\sum_{(x,y)\in \mathcal{D}_i}{\hspace{-10pt}\nabla \ell(f_{\hat{\theta}},y)} - \frac{1}{n_j}\sum_{(x,y)\in \mathcal{D}_j}\hspace{-10pt}\nabla \ell(f_{\hat{\theta}},y) }^2\\
\approx & \norm{ \mathbb{E}_{z\sim P_i}\nabla \ell(f_{\hat{\theta}},y) - \mathbb{E}_{z\sim P_j}\nabla \ell(f_{\hat{\theta}},y) }^2
\label{eq3}
\end{align*}
where the quality of the approximation depends on the number of samples $n_i$ and $n_j$.  The mixing coefficients for user $i$ are then set to the following normalized exponential function
\begin{equation}
  w_{i,j}=\frac{\frac{n_j}{n_i}e^{-\frac{1}{2\sigma_i\sigma_j}\Delta_{i,j}(\hat{\theta}) }}{\sum^m_{j'=1}\frac{n_{j'}}{n_i}e^{-\frac{1}{2\sigma_i\sigma_{j'}}\Delta_{i,j'}(\hat{\theta}) }} \hspace{0.5cm}\textnormal{for $j=1,\dots,m$}.
  \label{eq4}
\end{equation}
The mixture coefficients are calculated at the PS during a special round prior to federated training. During this round, the PS broadcasts a common model denoted $\hat{\theta}$ to the users, which compute the full gradient on their local datasets. At the same time, each node $i$ locally estimates the value $\sigma^2_i$ partitioning the local data in $K$ batches $\{\mathcal{D}^k_i\}^K_{k=1}$ of size $n_k$ and computing
\begin{equation} 
\sigma^2_i = \frac{1}{K}\sum_{k=1}^K\norm{\frac{1}{n_k} \sum_{(x,y)\in \mathcal{D}^k_i}\hspace{-10pt}\nabla \ell(f_{\hat{\theta}},y)-\frac{1}{n_i}\sum_{(x,y)\in \mathcal{D}_i}\hspace{-10pt}\nabla \ell(f_{\hat{\theta}},y )}^2   \label{eq9}
\end{equation} 
where $\sigma^2_i$ is an estimate of the gradient variance computed over local datasets $\mathcal{D}^k_i$ sampled from the same target distribution. Once all the necessary quantities are computed, they are uploaded to the PS, which proceeds to calculate the mixture coefficients and initiates the federated training using the custom aggregation scheme given by (\ref{eq3}).
Note that the proposed heuristic embodies the intuition provided by Th. \ref{Th1}. In fact, in the case of homogeneous users, it falls back to the standard FedAvg aggregation rule, while in the case of node $i$ has an infinite amount of data it degenerates to the local learning rule which is optimal in that case.
\subsection{Reducing the communication load}
A full-fledged personalization by the means of the user-centric aggregation rule (\ref{eq3}) would introduce a $m$-fold increase in communication load during the downlink phase as the original broadcast transmission is replaced by unicast ones. Although from a learning perspective the user-centric learning scheme is beneficial, it is also possible to consider overall system performance from a learning-communication trade-off point of view. The intuition is that, for small discrepancies between the user data distributions, the same model transfer positively to statistically similar devices. In order to strike a suitable trade-off between learning accuracy and communication overhead we hereby propose to adaptively limit the number of personalized downlink streams. In particular, for a number of personalized models $m_t$, we run a $k$-means clustering scheme with $k=m_t$ over the set of collaboration vectors $\{w_i\}_{i=1}^m$ and we select the centroids $\{\hat{w}_i\}_{i=1}^{m_t}$ to implement the $m_t$ personalized streams. We then proceed to replace the unicast transmission with group broadcast ones, in which all users belonging to the same cluster $c$ receive the same personalized model $\hat{w}_c$. Choosing the right value for the number of personalized streams is critical in order to save communication bandwidth but at the same time obtain satisfactory personalization capabilities. It can be experimentally shown that clustering quality indicators such as the Silhouette score over the user-centric weights can be used to guide the search for the suitable number of streams $m_t$ \cite{Journal}.
\section{Experiments}
\begin{figure*}
\centering
    \begin{subfigure}[t]{0.29\textwidth}
         \centering
         \includegraphics[width=\textwidth]{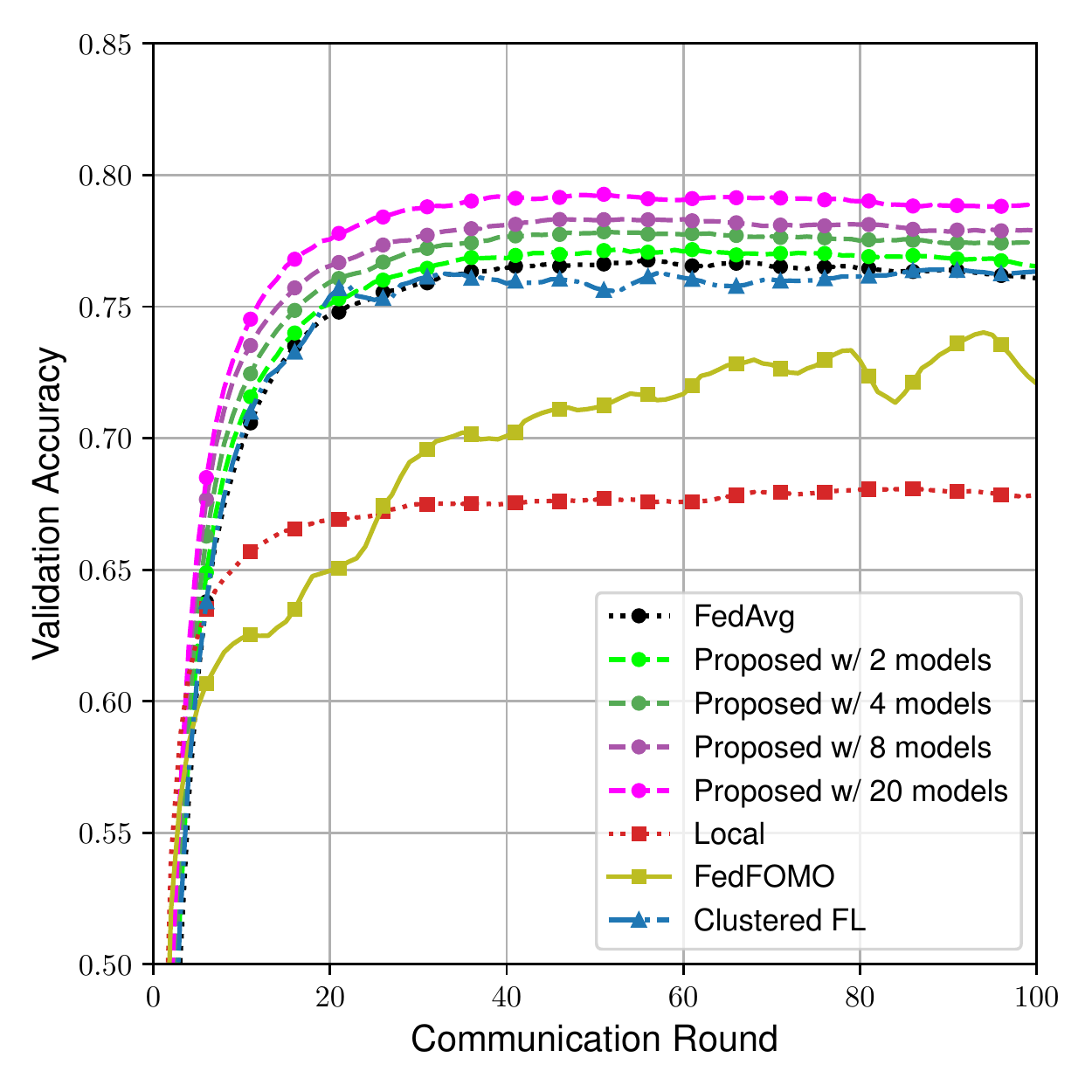}
         \centering
         \caption{EMNIST + label shift}
         \label{fig:EMNIST_label}
     \end{subfigure}
      \begin{subfigure}[t]{0.29\textwidth}
         \centering
         \includegraphics[width=\textwidth]{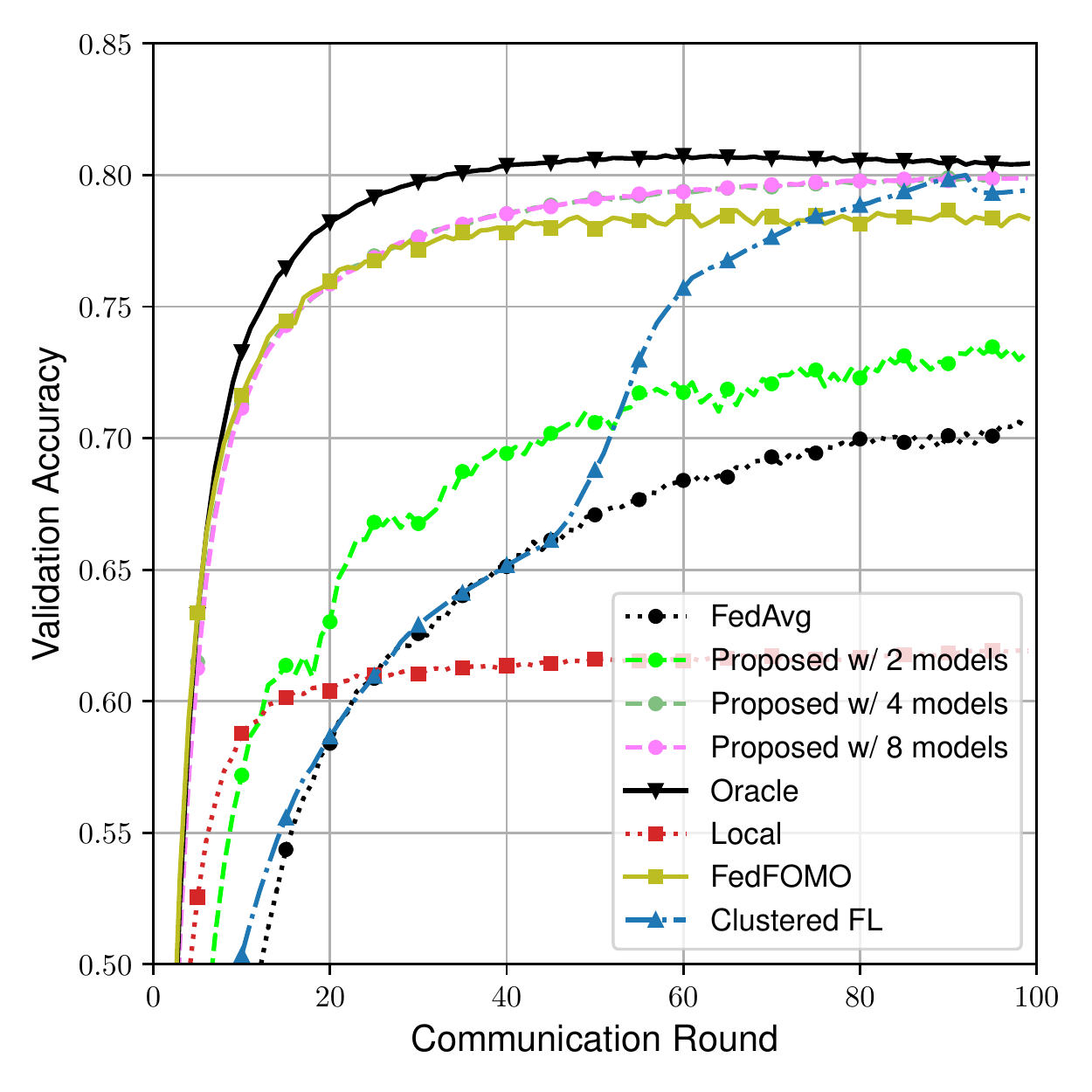}
         \caption{EMNIST + label and covariate shift}
         \label{fig:EMNIST_cov}
     \end{subfigure}
     \begin{subfigure}[t]{0.29\textwidth}
         \centering
          \includegraphics[width=\textwidth]{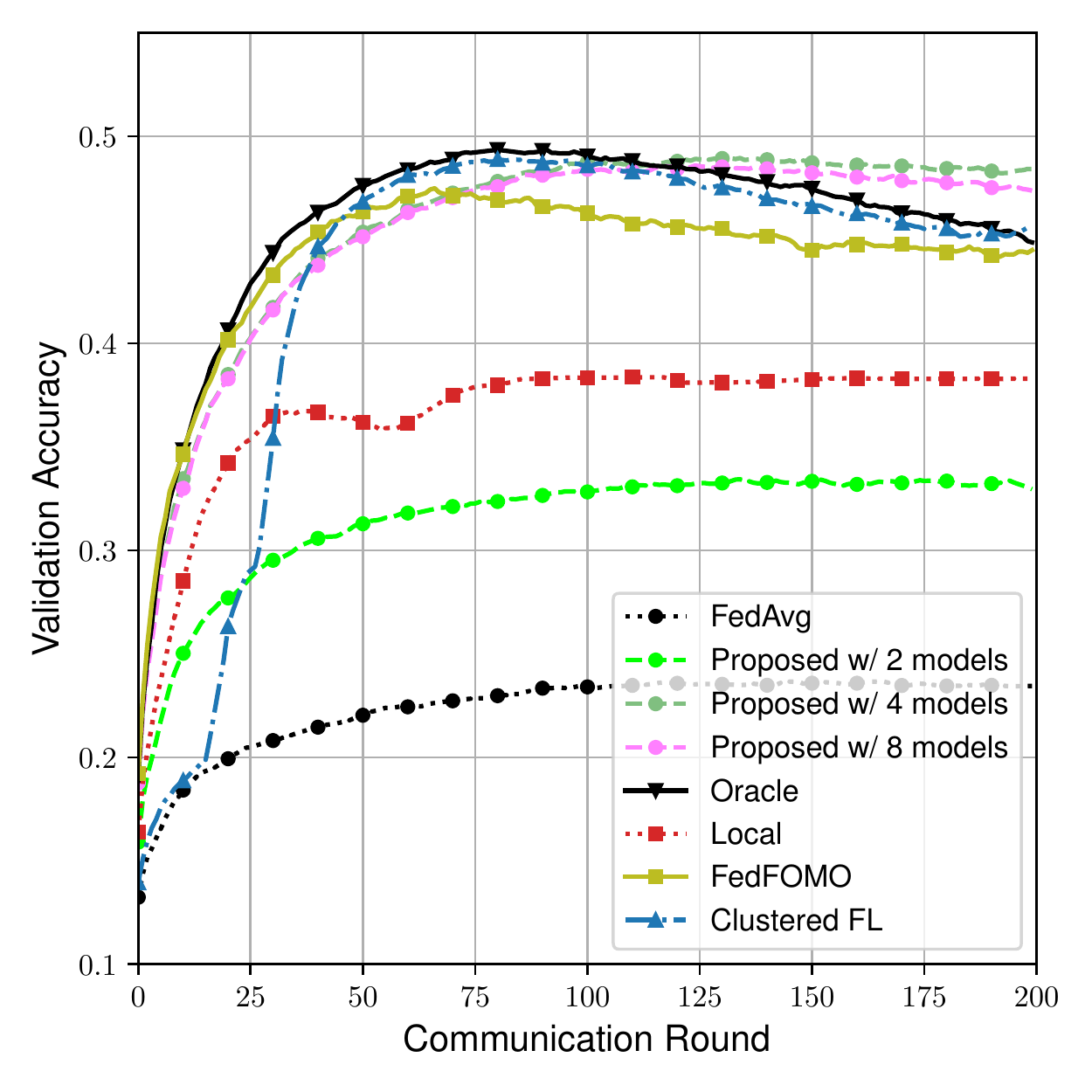}
         \caption{CIFAR10 + concept shift}
         \label{fig:CIFAR}
     \end{subfigure}
     \caption{Evolution of the average validation accuracy in the three simulation scenarios.}
\end{figure*}
We now provide a series of experiments to showcase the personalization capabilities and communication efficiency of the proposed algorithm.
\subsection{Set-up}
In our simulation we consider a handwritten character/digit recognition task using the EMNIST dataset \cite{cohen2017emnist}  and an image classification task using the CIFAR-10 dataset \cite{krizhevsky2009learning}.
Data heterogeneity is induced by splitting and transforming the dataset in a different fashion across the group of devices. In particular, we analyze three different scenarios:
\begin{itemize}
    \item\textbf{Character/digit recognition with user-dependent label shift} in which 10k EMNIST data points are split across 20 users according to their labels. The label distribution follows a Dirichlet distribution with parameter 0.4, as in \cite{mixture2021s,wang2020tackling}.
    \item\textbf{Character/digit recognition with user-dependent label shift and covariate shift} in which 100k samples from the EMNIST dataset are partitioned across 100 users each with a different label distribution, as in the previous scenario. Additionally, users are clustered in 4 group and at each group images are rotated of $\{0^{\circ},90^{\circ},180^{\circ},270^{\circ}\}$ respectively. 
     \item\textbf{Image classification with user-dependent concept shift} in which the CIFAR-10 dataset is distributed across 20 users which are grouped in 4 clusters, for each group we apply a different random label permutation. 
\end{itemize}
For each scenario, we aim at solving the task at hand by leveraging the distributed and heterogeneous datasets. We compare our algorithm against four different baselines: FedAvg, local learning, CFL \cite{sattler2020clustered} and FedFomo \cite{zhang2020personalized}.
In all scenarios and for all algorithms, we train a LeNet-5 convolutional neural network \cite{lecun1998gradient} using a stochastic gradient descent optimizer with a fixed learning rate $\eta=0.1$ and momentum $\beta=0.9$.
%The first algorithm is the standard implementation and it is known to perform well in case of i.i.d. data. local learning is instead a valid approach in case the severe data heterogeneity across users renders aggregation of data detrimental.  CLF approach falls in-between of the two, being able to detect data heterogeneity and reacting to it by clustering users in homogeneous groups during training. The latter approach, as ours, results in a personalized unique model for each user.

\begin{table*}[]
\centering
\caption{Worst user performance averaged over 5 experiments.}
\begin{tabular}{@{}ccccccc@{}}
\toprule
                            & Local & FedAvg & Oracle             & CFL \cite{sattler2020clustered} & FedFOMO \cite{zhang2020personalized}      & Proposed           \\ \midrule
EMNIST label shift          & 58.8  & 68.9   & -                  & 70.3   &      70.0     & \textbf{73.2} for $k=20$    \\
EMNIST covariate and label shift & 56.0    & 67.5     & 77.4 & 76.1 &    73.6  & \textbf{76.4} for $k=4$               \\
CIFAR concept shift         &35.7     & 19.6     & 49.1 & 48.6 &   45.5   & \textbf{49.1} for $k=4$               \\ \bottomrule
\end{tabular}

\label{table:worst}
\end{table*}
\subsection{Personalization performance}
We now report the average accuracy over 5 trials attained by the different approaches. We also study the personalization performance of our algorithm when we restrain the overall number of personalized streams, namely the number of personalized models that are concurrently learned. In Fig.\ref{fig:EMNIST_label} we report the average validation accuracy in the EMNIST label shift scenario. We first notice that in the case of label shift, harvesting intelligence from the datasets of other users amounts to a large performance gain compared to the localized learning strategy. This indicates that data heterogeneity is moderate and collaboration is fruitful.  Nonetheless, personalization can still provide gains compared to FedAvg. Our solution yields a validation accuracy which is increasing in the number of personalized streams. Allowing maximum personalization, namely a different model at each user, we obtain a 3\% gain in the average accuracy compared to FedAvg. CFL is not able to transfer intelligence among different groups of users and attains performance similar to the FedAvg. This behavior showcases the importance of soft clustering compared to the hard one for the task at hand. We find that FedFOMO, despite excelling in the case of strong statistical heterogeneity, fails to harvest intelligence in the label shift scenario. In Fig.\ref{fig:EMNIST_cov} we report the personalization performance for the second scenario. In this case, we also consider the oracle baseline, which corresponds to running 4 different FedAvg instances, one for each cluster of users, as if the 4 groups of users were known beforehand.  Different from the previous scenario, the additional shift in the covariate space renders personalization necessary in order to attain satisfactory performance. In fact, the oracle training largely outperforms FedAvg. Furthermore, as expected, our algorithm matches the oracle final performance when the number of personalized streams is 4 or more. Also, CLF and FedFOMO are able to correctly identify the 4 clusters. However, the former exhibits slower convergence due to the hierarchical clustering over time while the latter plateaus to a lower average accuracy level. We turn now to the more challenging CIFAR-10 image classification task. In Fig.\ref{fig:CIFAR} we report the average accuracy of the proposed solution for a varying number of personalized streams, the baselines, and the oracle solution. As expected, the label permutation renders collaboration extremely detrimental as the different learning tasks are conflicting. As a result, local learning provides better accuracy than FedAvg. On the other hand, personalization can still leverage data among clusters and provide gains also in this case. Our algorithm matches the oracle performance for a suitable number of personalized streams. This scenario is particularly suitable for hard clustering, which isolates conflicting data distributions. As a result, CFL matches the proposed solution. FedFOMO promptly detects clusters and therefore quickly converges, but it attains lower average accuracy compared to the proposed solution. 

The performance reported so far is averaged over users and therefore fails to capture the existence of outliers performing worse than average. In order to assess the fairness of the training procedure, in Table \ref{table:worst} we report the worst user performance in the federated system. The proposed approach produces models with the highest worst case in all three scenarios.

% \usepackage{booktabs}
% \usepackage{booktabs}

\begin{figure*}
\centering
    \begin{subfigure}[t]{0.29\textwidth}
         \centering
         \includegraphics[width=\textwidth]{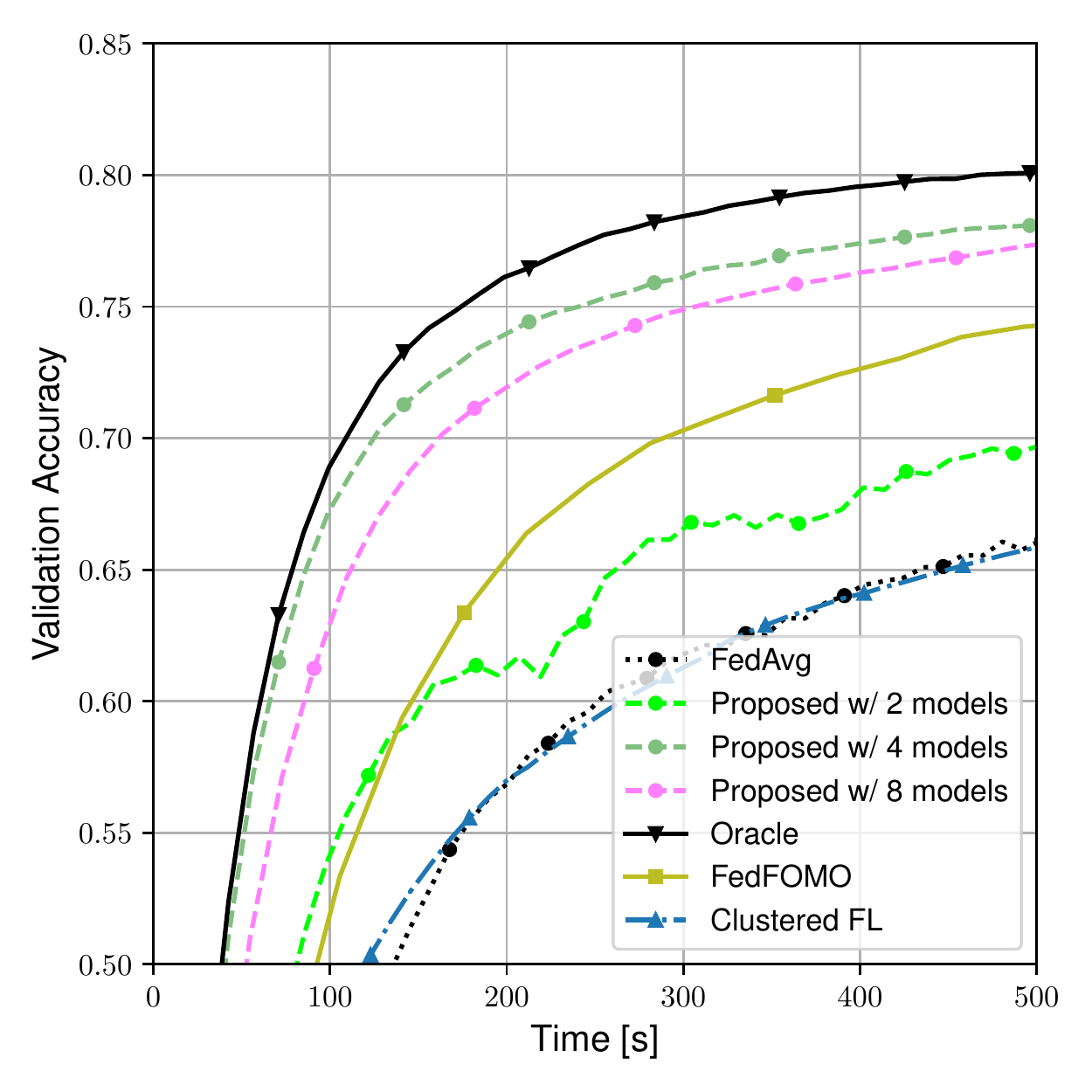}
         \centering
         \caption{$\rho=4, T_{min}=T_{dl}=\frac{1}{\mu}$}
         \label{fig:1}
     \end{subfigure}
      \begin{subfigure}[t]{0.29\textwidth}
         \centering
         \includegraphics[width=\textwidth]{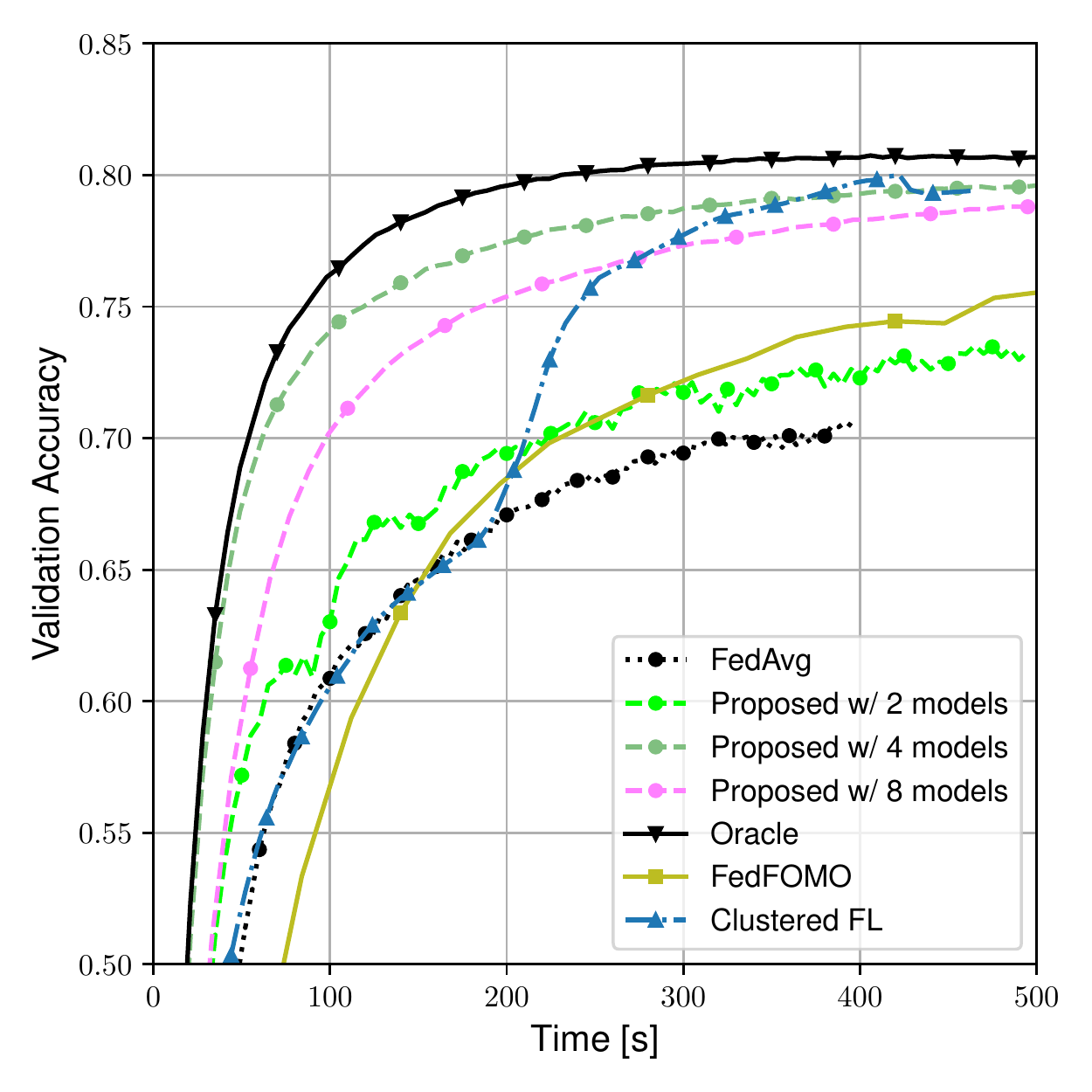}
         \caption{$\rho=2, T_{min}=T_{dl},\frac{1}{\mu}=0$}
         \label{fig:2}
     \end{subfigure}
     \begin{subfigure}[t]{0.29\textwidth}
         \centering
          \includegraphics[width=\textwidth]{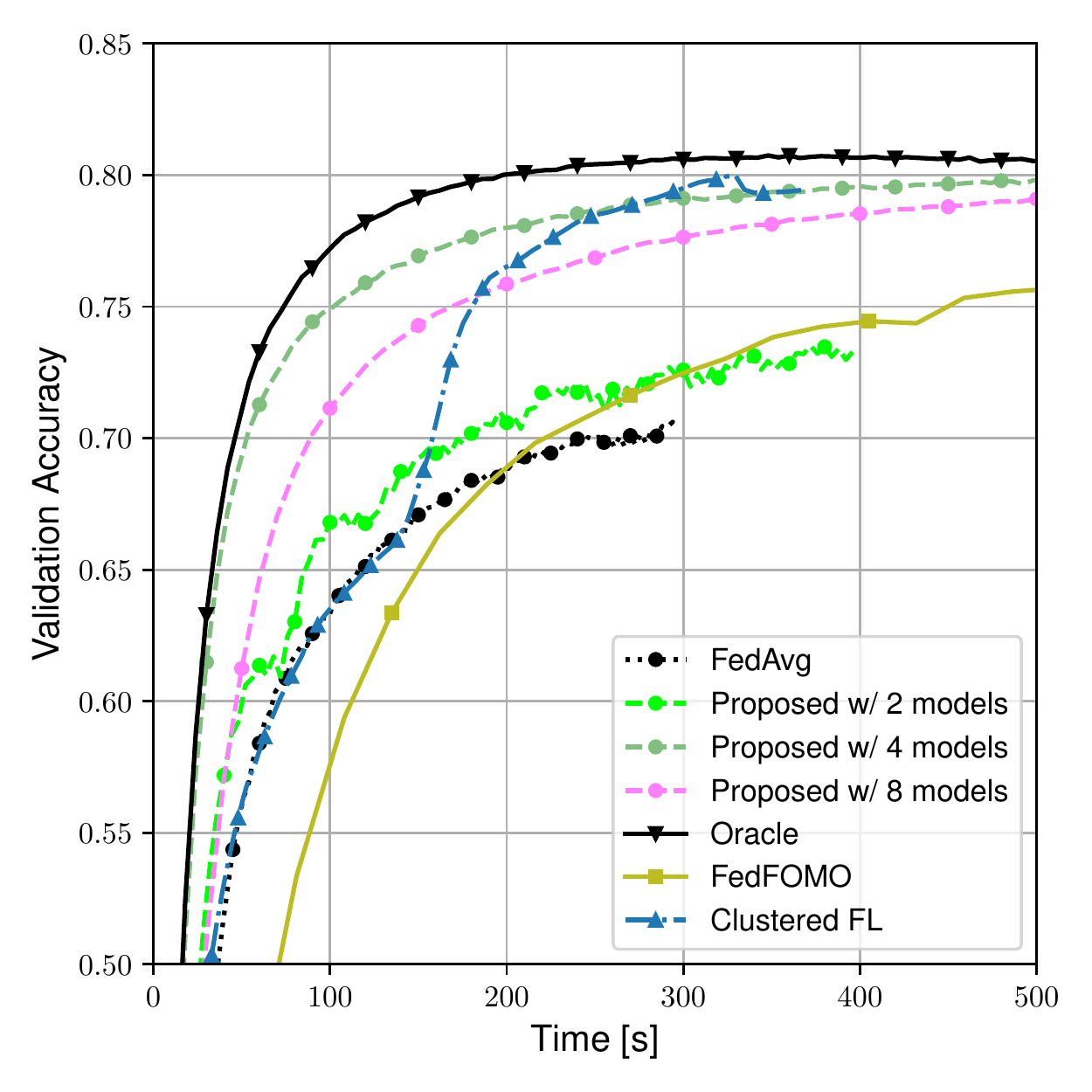}
         \caption{$\rho=1, T_{min}=T_{dl},\frac{1}{\mu}=0$}
         \label{fig:3}
     \end{subfigure}
     
     \caption{Evolution of the average validation accuracy against time normalized w.r.t. $T_{dl}$ for the three different systems.}
     \label{fig:comm_eff}
\end{figure*}

\subsection{Communication Efficiency}
Personalization comes at the cost of increased communication load in the downlink transmission from the PS to the federated user. In order to compare the algorithm convergence time, we parametrize the distributed system using two parameters. We define by $\rho=\frac{T_{ul}}{T_{dl}}$ the ratio between model transmission time in uplink (UL) and downlink (DL). Typical values of $\rho$ in wireless communication systems are in the $[2,4]$ range because of the larger transmitting power of the base station compared to the edge devices. Furthermore, to account for unreliable computing devices, we model the random computing time $T_i$ at each user $i$ by a shifted exponential r.v.  with a cumulative distribution function
\[
P[T_i>t]=1-\mathbbm{1}(t\geq T_{min})\left[1-e^{-\mu(t-T_{min})}\right]
\]
where $T_{min}$ representing the minimum possible computing time and $1/\mu$ being the average additional delay due to random computation impairments. Therefore, for a population of $m$ devices, we then have 
\[
T_{comp}=\mathbb{E}\left[\max\{T_1,\dots,T_m\}\right]=T_{min}+\frac{H_m}{\mu}
\]
where $H_m$ is the $m$-th harmonic number.
To study the communication efficiency we consider the simulation scenario with the EMNIST dataset with label and covariate shift.  
In Fig. \ref{fig:comm_eff} we report the time evolution of the validation accuracy in 3 different systems. A wireless systems with slow UL $\rho=4$ and unreliable nodes $T_{min}=T_{dl}=\frac{1}{\mu}$, a wireless system with fast uplink $\rho=2$ and reliable nodes $T_{min}=T_{dl}$, $\frac{1}{\mu}=0$ and a wired system $\rho=1$ (symmetric UL and DL) with reliable nodes $T_{min}=T_{dl}$, $\frac{1}{\mu}=0$. The increased DL cost is negligible for wireless systems with strongly asymmetric UL/DL rates and in these cases, the proposed approach largely outperforms the baselines. In the case of more balanced UL and DL transmission times $\rho=[1,2]$ and reliable nodes, it becomes instead necessary to properly choose the number of personalized streams in order to render the solution practical. Nonetheless, the proposed approach remains the best even in this case for $k=4$. Note that FedFOMO incurs a large communication cost as personalized aggregation is performed at the client-side.
\section{Conclusion}In this work, we presented a novel federated learning algorithm that exploits multiple user-centric aggregation rules to produce personalized models. The aggregation rules are based on user-specific mixture coefficients that can be computed during one communication round prior to federated training. Additionally, in order to limit the communication burden of personalization, we propose a simple strategy to effectively limit the number of personalized streams. We experimentally study the performance of the proposed solution across different tasks. Overall, our solution yields personalized models with higher testing accuracy while at the same time being more communication-efficient compared to the competing baselines.

\bibliographystyle{unsrt}
\bibliography{conference_101719}
% \begin{algorithm}[h]
% \SetAlgoLined
% \DontPrintSemicolon
 
% \SetKwInOut{Input}{Input}
% \SetKwInOut{Output}{Output}
%   \Input{$K$ nodes indexed by $k$, $B$ is the local mini-batch size, $E$ is the number of Epochs, $\eta$ is the learning rate and the data set at each node $\{\mathcal{D}_k\}^K_{k=1}$ .}
%   \Output{$\theta^* = \argmin_{\theta}L(\theta)$\newline}

%   \textnormal{\textbf{Orchestrator Executes:}}  \;
%     {\textnormal{Initialize $\theta_0$}\\
%             \For{\textnormal{each round} $t = 0,1,...,T$}
%                 {   \textnormal{Orchestrator Broadcasts} $\theta^t$ \;
%             \For{each node $\textbf{k}$ in Parallel}
%             {
%                     $\theta^{t+1}_k\gets \textnormal{\textbf{ClientUpdate}}(\theta^t,\mathcal{D}_k)$
%             }
%          $\theta^{t+1}\gets \frac{1}{\sum^K_{i=1}|\mathcal{D}_i|}\sum^K_{k=1}|\mathcal{D}_k| \theta^{t+1}_k$ 
%                 }}
%     $ \textnormal{\textbf{ClientUpdate}}(\theta,\mathcal{D}_k): \hspace{0.4cm} //\textbf{\textit{Run on node $k$}}$   
%     {
%     \; $\mathcal{B}$ $\gets$ $\textnormal{Split}$ $\mathcal{D}_k$ $\textnormal{into batches of size $B$}$
%          \;\For{\textnormal{each local Epoch $i$ from} $ 1\rightarrow E$}
%             { \For{\textnormal{batch} $b\in \mathcal{B}$}
%                 {   $\theta\gets\theta - \eta \nabla \ell(\theta,b)$
%                 }
        
%             }
%     return $\theta$ to \textit{Orchestrator}
%     }

% \caption{Federated Averaging}
% \label{Algo0} 
% \end{algorithm} 

\end{document}